\pdfoutput=1

\documentclass[11pt]{article}

\usepackage{acl}

\usepackage{times}
\usepackage{latexsym}

\usepackage[T1]{fontenc}

\usepackage[utf8]{inputenc}

\usepackage{microtype}

\usepackage{inconsolata}

\usepackage{graphicx}
\usepackage{amsmath}
\usepackage{algorithm} 
\usepackage{algpseudocode}
\usepackage{todonotes}
\usepackage{soul}
\usepackage{multirow} 
\usepackage{booktabs}  
\usepackage{tabularx}
\usepackage{listings}
\algrenewcommand\algorithmicrequire{\textbf{Input:}}
\algrenewcommand\algorithmicensure{\textbf{Output:}}
\title{THELMA: Task Based Holistic Evaluation of Large Language Model Applications-RAG Question Answering}

\author{
 \textbf{Udita Patel}\thanks{Equal contribution.},
 \textbf{Rutu Mulkar}\textsuperscript{*},
 \textbf{Jay Roberts}\textsuperscript{*}\thanks{Work done at Amazon.},
 \textbf{Cibi Chakravarthy Senthilkumar\textsuperscript{*}},
\\
 \textbf{Sujay Gandhi},
 \textbf{Xiaofei Zheng},
 \textbf{Naumaan Nayyar},
 \textbf{Parul Kalra},
 \textbf{Rafael Castrillo}
\\
\\
Amazon.com Services Inc.
\\
\small{ \texttt{ \{patudita, rmulkar, jarberts, sentocb, sujaygan, 
zhexiaof, nayyarnn, kalparul, rbarquer}\}@amazon.com}}

\begin{document}

\maketitle
\begin{abstract}
We propose THELMA (\textbf{T}ask Based \textbf{H}olistic \textbf{E}valuation of Large \textbf{L}anguage \textbf{M}odel \textbf{A}pplications), a reference free framework for RAG (Retrieval Augmented generation) based question answering (QA) applications. THELMA consist of six interdependent metrics specifically designed for holistic, fine grained evaluation of RAG QA applications. THELMA framework helps developers and application owners evaluate, monitor and improve end to end RAG QA pipelines without requiring labelled sources or reference responses.We also present our findings on the interplay of the proposed THELMA metrics, which can be interpreted to identify the specific RAG component needing improvement in QA applications.
\end{abstract}
\section{Introduction}

Retrieval Augmented Generation (RAG) paradigm \citep{NEURIPS2020_6b493230} improves the performance of Large Language Models (LLMs) by using excerpts of external knowledge to generate responses. It consists of two fundamental components: a retriever and a generator. In RAG based question answering (QA), a retriever searches and retrieves sources related to the asked query; followed by feeding the query and the sources to a generator model(LLMs) tasked with generation of the response. Given that RAG enables the incorporation of external knowledge, they present an excellent choice for domain adaptation of QA application in enterprise applications. RAG based QA has been widely used in automation of customer care services \citep{10.1145/3626772.3661370, zhang-etal-2024-rag4itops} operations and as an enhanced capability within conventional information retrieval applications. Once developed and deployed, these applications require comprehensive monitoring and granular, instance level error diagnosis for performance evaluation as well as continual improvement.
Evaluating and improving retrievers and generators  components individually don't necessarily have effect on overall performance of downstream RAG QA applications. Additionally, performance evaluation of enterprise QA applications present several unique challenges: 

\begin{enumerate}
    \item Unavailability of reference responses: Enterprise QA applications are typically built for a domain specific source corpora where these sources are continually updated, and expanded with new sources. Continual changes in the source corpora render the reference responses and/or its annotation obsolete.

     \item Granular Error Diagnosis: Monitoring of production traffic on QA applications requires analyzing erroneous  interactions in fine grained, interpretable manner. This allows for targeted improvements to the various components of QA applications. Existing metrics like relevancy does not provide interpretable scores to improve a specific component in RAG systems.

    \item Varying tolerance levels for sub-optimal responses: Based on the criticality of use case, the tolerance towards inaccuracies varies across different dimensions of QA applications. This poses a requirement on evaluation framework to accommodate adaptable threshold mechanisms across dimensions. 

    \item  Holistic evaluation for robust applications: During development and iterative optimization of retrievers and generators, improvements targeted at specific component can inadvertently degrade performance of another. An evaluation framework should enable assessment of individual components as well as their interplay in order to maintain the robustness of the application.
\end{enumerate} 
To overcome these challenges, we propose THELMA as a reference free, multi-dimensional, granular, domain agnostic, interpretable framework for evaluation of RAG QA applications. THELMA utilizes query, source and response as input, and processes it into logical decompositions for granular assessment. The decompositions of query, source and response are then matched against other two inputs to obtain metrics score. Our overall contribution are as follows:

\begin{enumerate}
\item We have formally defined and implemented THELMA, a reference free evaluation framework that identifies specific areas for improvement in the implementation of RAG-based applications through its metric interplay.
\item We have redefined the notion of response irrelevance at a finer level of granularity with three distinct metrics - evaluating non-essentiality with precision, repetition with distinctness, and incorrectness with groundedness. 
\item We demonstrate metrics reliability by comparing agreement of RAGAs and THELMA metrics with preferential human annotation.  
\end{enumerate}


%
%
\section{Related work}
\label{sec:related-work}
Traditional metrics for evaluation of retrievers like recall@k and MRR \citep{voorhees-tice-2000-trec} depends on annotation of sources and do not capture the semantic scope of knowledge base. Similarly, generator metrics like BERTScore \citep{zhang2020bertscoreevaluatingtextgeneration} require ground truth references as well as aren't interpretable. 
In recent years, large language models have also been used as evaluators for various natural language processing (NLP) tasks under the LLM-as-a-judge paradigm. \citet{10.5555/3666122.3668142} examined the usage and limitation of chatbots and question answering(QA) for open ended domain, but enterprise QA applications are typically closed domain. \citet{ru2024ragcheckerfinegrainedframeworkdiagnosing} proposed RAGchecker with granular metrics which are similar to the ones proposed in THELMA but are not suitable for monitoring of real world applications where references responses are unavailable. Another class of frameworks focuses on scenarios when reference responses and ground-truth annotation for retriever are unavailable. \citet{es-etal-2024-ragas} proposed RAGAs with three key dimensions as introduced in Trulens \citep{trulensTriad}: context relevance, answer relevance and faithfulness. ARES \citep{saad-falcon-etal-2024-ares} evaluated along the same dimension as RAGAs, but curated a small set of human annotations and uses fine tuned LLMs as judges. However, relevance measured in these metrics is coarse grained and cannot provide actionable diagnosis needed for continuous improvements of applications. \citet{wang2024evaluatingqualityanswersretrievalaugmented} employs a rubric based approach to evaluate responses where the LLM also produces reasoning along with the scores but does not address retriever evaluation.
Model based evaluations have also been proposed for diagnosis of specific issues observed in RAG based question answering. Incorrectness in responses identified by validating against a reference source has been evaluated by FactScore \citep{min-etal-2023-factscore}  and Refchecker \citep{hu_refchecker_2024}. We include FactScore's methodology termed as groundedness in our framework. Presence of irrelevant and absence of a relevant source in a set of retrieved sources has also been evaluated previously. \citep{salemi2024evaluating} introduced eRAG where sources in a ranked retrieved set are evaluated for relevancy based on the  ground truth labels of the RAG downstream task. 
RAGChecker \citep{ru_ragchecker_2024} measures context precision and recall by comparing sources with reference answers, while THELMA measures source precision by evaluating presence of a irrelevant source and absence of an essential source w.r.t to the question under consideration making it reference free. We establish similar precision-coverage trade-off with response metrics. \citet{zhang2024verbosityneqveracitydemystify} evaluates repetition verbosity of responses with performance difference and relative performance difference metrics but requires reference responses, unlike THELMA's self distinctness metric.  

\begin{figure*}
    \centering
    \makebox[\textwidth]{\includegraphics[width=\textwidth]{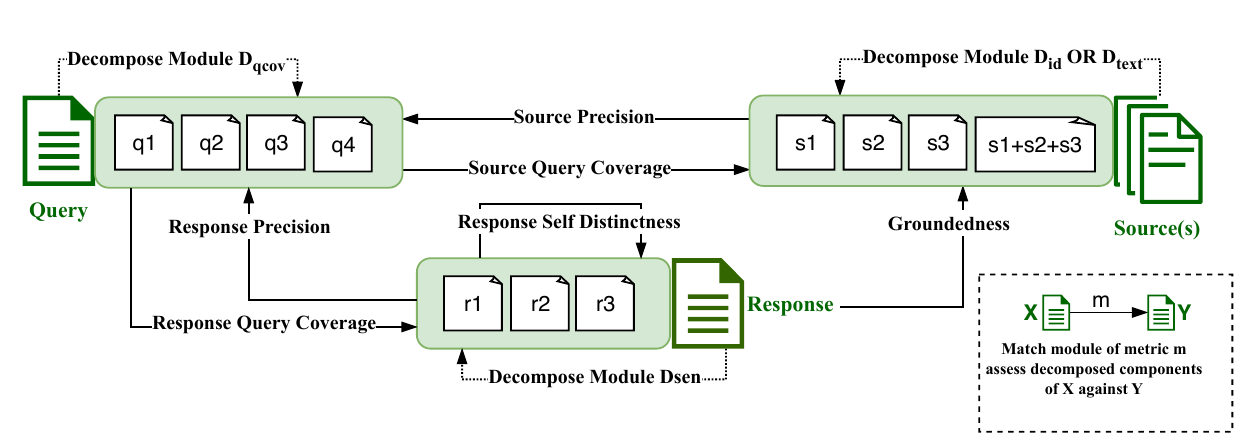}}
    \caption{Overview of THELMA framework with RAG triad components; Query, Source(s) and Response.}
    \label{fig:enter-label}
\end{figure*}

\section{THELMA Framework}\label{framework}
THELMA is a suite of LLM-as-Judge metrics to evaluate RAG based QA applications. Proposed metrics are based on RAG triad but enables more granular diagnosis. RAG Triad is an evaluation framework introduced by trulens \citep{trulensTriad} to holistically assess reliability of LLM generated responses in a QA application. It consists of query($q$), retrieved sources($s$) and response($r$) generated by the LLM, evaluated using context relevance, response relevance and groundedness using model based evaluations.

We propose our evaluation framework under the following assumptions: 1) Query, source, response triplets are available 2) Generator should not respond with internal knowledge; the source corpora contains only accurate information, hence can be treated as ground truth. 3) Reference responses are unavailable. Each metric is defined with three modules; Decompose, Match and Aggregate. 
\begin{enumerate}
\item Decompose module fragments the textual inputs into independent logical components, hereon referred to as \textit{decomposed components}. Implementation is similar to claim extraction in \citet{min-etal-2023-factscore}.      
\item Match module assesses a decomposed component against other textual inputs based on metric specific criterion.
\item Aggregate module combines individual matching scores on decomposed components to produce a score for a specific metric on given (query, source response) triplet.
\end{enumerate}
We demonstrate the reliability of proposed metrics when compared against human judgement in table \ref{tab:response-metrics-results} and effectiveness of framework in interpreting actionable insights in table \ref{tab:interpretation}.





\section{THELMA Metrics}
Formally, let $G$ be the function representing a generator LLM. Let $T$ be the retriever model used to retrieve set $s$ from $S$ using $q$ as input. $G$ generates response $r\in R$ for a query $q\in Q$ using retrieved source set $s\in S$. 
\begin{equation}
s=T(q), r=G(q,s)
\end{equation}
$G$ and $T$ are assumed to be inaccessible for the purposes of all metrics calculations. All metrics follow the same set of steps. First, a decompose module $D:X\mapsto \{x_1,..x_n\}$ transforms given textual input into independent atomic components. Next, a match module $m:X\times Y \mapsto \{0,1\}$ which assesses two textual inputs based on the metric specific criterion. Match module outputs 1 if the criteria is satisfied, 0 otherwise. Lastly, an aggregate modules calculates a metric score over an instance of $(q,s,r)$ triad  as the macro average of the matching function, unless specified otherwise. The choice of decompose and match module, textual input and the aggregate function defines each metric. All metrics produce scores between 0 and 1 using combination of two text inputs from the $(q, s, r)$ triplet, with higher scores denoting better performance.
\subsection{Source Precision}
\label{sp}
Retrievers sometimes can retrieve related but unnecessary sources for a given query \citep{wu2024how}. These irrelevant sources progressively degrade response quality by detracting LLMs from generating correct answers \citep{10.1145/3626772.3657834}. Presence of unnecessary sources is evaluated with source precision.

Source precision is the degree to which the retrieved source set $s$ is essential to answer the query. Source precision of set $s$ with respect to $q$, is
\begin{equation}
\begin{split}
    \text Source\_Precision(s, q)\\
    = \frac{1}{|D_{id}(s)|}\sum_{s_j\in D_{id}(s)} m_{sp1}(s_j, q).
\end{split} 
\end{equation}
The necessary and unnecessary components in a source set can be evaluated with two variants. First variant is as follows. \textbf{Decompose module}: $D_{id}$ is an identity function, hence each source $s_j$ remains unchanged as retrieved by the retriever. \textbf{Match module}: $m_{sp1}$ operates at a single source level, assigning a score of $1$ to entire source $s_j$ that contains any essential information for answering the query. This diagnosis can uncover issues in retriever, where a retrieved source $s_j$ is entirely non-essential for answering any query.
\begin{equation}
\begin{split}
    \text Source\_Precision(s, q)\\
    = \frac{1}{|D_{text}(s)|}\sum_{s_j\in D_{text}(s)} m_{sp2}(s_j, q)
\end{split} 
\end{equation}
The second variant, aims to identify all non essential facts within the entire source set $s$. \textbf{Decompose module}: $D_{text}$ transforms the source set $s$ into a set of components, where each component $s_j$ is a standalone fact, conveying one piece of information. \textbf{Match module}: $m_{sp2}$ assesses the essentiality of a decomposed source component for answering the query. This granular level evaluation of $s$ quantifies the presence of total amount of non-essential information. This diagnosis can uncover issues in source chunking strategies as well as in how the presence of non essential information in the sources is detracting the LLM from generating the correct answer. 
\subsection{Source Query Coverage}\label{sqc}
Retrievers will sometimes fail to retrieve sources containing the answer(s) for all or few of sub-queries. Ideally, generator should reject answering such queries (or part thereof ) but only do so with $45\%$ success rate, and instead use its intrinsic knowledge to generate ungrounded responses \citep{chen2023benchmarkinglargelanguagemodels}. Therefore, absence of necessary sources required to fully answer given query is evaluated with source query coverage.

Source query coverage is degree to which a source set $s$ answers all components of the decomposed query. Source query coverage of $s$ with respect to $q$, is
\begin{equation}
\begin{split}
\text Source\_Query\_Coverage(s, q) = \\
      \frac{1}{|D_{qcov}(q)|}\sum_{q_l\in D_{qcov}(q)} \max_{s_j\in D_{ic}(s)}m_{sqcov}(q_l, s_j).
\end{split}
\end{equation}
\textbf{Decompose module}: $D_{qcov}$ transforms query $q$ into a set of decomposed query components, where each component represents a standalone question from the original query while resolving pronouns, ignoring greeting and factual statements that do not represents a question.\textbf{Match module}: Query component $q_l$ is assessed independently for source components and also against the entire source set. Assessing with source components maintains prompt performance which was observed to deteriorate when the  number of retrieved sources or the length of the sources increased. Assessment with entire source set handles multi-hop query scenarios. $m_{sqcov}$ assesses the presence of an answer for query component $q_l$ in each of the source components $s_j\in D_{ic}(s)$. If any of source component answers $q_l$, the matching score is set as $1$.

\begin{table*}
\centering
\begin{tabular}{cll ccc ccc c}
    \toprule
\multirow{3}{*}{} 
        & & \multicolumn{3}{c}{Context Relevance} & \multicolumn{3}{c}{Answer Relevance} & Faith. \\
    \cmidrule(lr){3-5} \cmidrule(lr){6-8} \cmidrule(lr){9-9} 
        & Model & SP1  & SP2  & SQC  & RP  & RQC  & SD & GR \\
    \midrule
\multirow{3}{*}{RAGAs}  
        & Sonnet 3.5  & 0.90  & 0.35    & 0.70    & 0.64    & 0.50    & 0.26    & 0.96          \\
        & Haiku 3.5   & 0.94    & 0.42    & 0.17    & 0.26    & 0.43    & 0.42     & 0.50       \\
        & Llama 3.3 70b   & 0.92    &0.28    & 0.29    & 0.63    & 0.68    & 0.52    & 0.94        \\
    \addlinespace
\multirow{4}{*}{THELMA}
        & Sonnet 3.5  & \textbf{1.0}  & 0.35   & 0.76    & 0.84    & 0.75    & \textbf{1.0}    & 0.96         \\
        & Haiku 3.5   & \textbf{1.0}    & 0.42    &\textbf{0.82}     & \textbf{0.95}    & 0.87    & \textbf{1.0}    & 0.96       \\
        & Llama 3.3 70b   & \textbf{1.0}    &\textbf{0.64}    & 0.70     & 0.89    &\textbf{0.88}    & \textbf{1.0}    &\textbf{1.0}   \\
    \bottomrule
\end{tabular}
\caption{Agreement with human annotator in pairwise comparisons. Source Precision, Source Query Coverage is compared with coarse grained Context Relevance in RAGAs. Response Precision, Response Query Coverage, and Response Self Distinctness is compared with coarse grained Answer Relevance in RAGAs. Groundedness is compared with Faithfulness in RAGAs.}
\label{tab:response-metrics-results}
\end{table*}
\subsection{Response Precision}\label{rqc}
LLMs tend to prefer generating lengthy, verbose responses adding extraneous details along with necessary information required to answer the query \citep{zhang2024verbosityneqveracitydemystify}. This verbosity in enterprise applications using proprietary models (LLMs) leads to increased cost, as they often employ token-based pricing models. Presence of \textit{extraneous, unnecessary} information is evaluated with response precision metric.

Response precision is the degree to which the amount of information provided in response necessary to answer the query. A precise response answers the question directly and avoids additional unnecessary information. Response precision of $r$ with respect to $q$, is
\begin{equation}
\begin{split}
    \text Response\_Precision(r,q)\\
    = \frac{1}{|D_{text}(r)|}\sum_{r_k\in D_{text}(r)} m_{rp}(r_k, q).
\end{split}
\end{equation}

\textbf{Decompose module}: We use the same decompose $D_{text}$ to transform both the source set $s$ into facts and the generated response $r$ into claims.  \textbf{Match module:}  $m_{rp}$ assesses  the essentiality of a response component $r_k$ for answering any part of the query.

\subsection{Response Query Coverage}
Generator LLMs may produce incomplete and/or insufficient answers due to ambiguous generator prompts or language complexities in retrieved sources. Additionally, LLMs can struggle to identify relevant information when it's present within lengthy source texts, thereby missing answering the query \citep{liu-etal-2024-lost}. \textit{Incompleteness} of response is evaluated with response query coverage. 

Response query coverage is degree to which the response answers the query. Response query coverage of $r$ with respect to $q$, is
\begin{equation}
\begin{split}
\text Response\_Query\_Coverage(r, q) \\
= \frac{1}{|D_{qcov}(q)|}\sum_{q_i\in D_{qcov}(q)} m_{rqcov}(q_i, r).
\end{split}
\end{equation} 
\textbf{Decompose module}: $D_{qcov}$ used in source query coverage, see \ref{sqc}. \textbf{Match module}: $m_{rqcov}$ validates that the intent of the decomposed query component $q_i$ is addressed in response. It also assess whether the logical structure of the response resolves the query.  

\subsection{Response Self-Distinctness}
LLMs also demonstrate verbosity by repeating information through paraphrasing \citep{briakou2024implicationsverbosellmoutputs}. In enterprise applications, it is crucial to identify repetitiveness as it can decrease user readability while also contributing to cost and latency. Presence of \textit{repetitive, redundant information} is evaluated with response self-distinctness. 

Response self distinctness is the degree to which components of the response are dissimilar from each other. A self-distinct response will not repeat information within itself. Response self distinctness of $r$, is
\begin{equation}
\begin{split}
\text Response\_Self\_Distinctness(r)\\
=\frac{1}{|D_{sen}(r)|}\sum_{r_j, r_u\in D_{sen}(r)}(1- m_{sd}(r_j, r_u)).
\end{split}
\end{equation}
\textbf{Decompose Module}: $D_{sen}$ splits the response $r$ into separate sentences using primary sentence terminators i.e. Periods, exclamation marks, and question marks. \textbf{Match Module}: Match module  calculates cosine similarity between two response sentences represented with titan-v1 embedding. $(1-m_{sd})$ 
is treated as distinctness score.


\subsection{Groundedness}
A response $r$ generated by an LLM $M$ may contain facts derived from its intrinsic knowledge acquired during pre-training \citep{Huang_2025}. The intrinsic knowledge may not necessarily align with facts and domain specific information present in source $s$ provided as input.  For evaluating a closed-domain enterprise application, the domain knowledge stored in the authoritative source(s) is assumed to be the ground truth. Thus, we measure the presence of \textit{incorrectness} of the response $r$ based on its infidelity to the facts present in provided source(s) $s$. Groundedness of $r$ with respect to $s$, is
\begin{equation}
\label{eq:groundedness}
\begin{split}
    \text{Groundedness}(r, s)\\
    = \frac{1}{|D_{text}(r)|}\sum_{r_i \in D(r)} m_{gr}(r_i, s).
\end{split}
\end{equation}
\textbf{Decompose Module}: $D_{text}$ transforms response $r$ in to a set of independent claims, where each component is a sentence representing a stand alone factual statement extracted from the response as proposed in \citet{min-etal-2023-factscore}. \textbf{Match Module}:  validates each fact component $r_i$ against the given source set $s$. Matching module takes as input a fact component $r_i$, as well as the retrieved source set $s$.

\section {Experiments}
\label{sec:experiments}
\begin{table*}
\centering%
\begin{tabularx}{\textwidth}{lXr}
\toprule
\textbf{Metric interplay} & \textbf{Interpretation}  & \textbf{Component to be improved} \\
\midrule
SD$\downarrow$ RP$\uparrow$& Lengthier responses with relevant but repetitive information, low user readability. & Prompt or Generator \\
\midrule
SQC$\downarrow$ RQC$\downarrow$ & Inaccurate retrieval OR missing information in source corpora. & Retriever or Source text. \\
\midrule
SP$\downarrow$ SQC$\uparrow$ & All components of the query are being addressed, but some retrieved sources are only loosely relevant. & Retriever \\
\midrule
RQC$\downarrow$ SQC$\uparrow$ & Information required to answer the query is present in source but not used in response.  & Prompt or Generator \\
\midrule
RP$\downarrow$ $SP_1$$\uparrow$ & Response contains extraneous information but majority of retrieved sources are essential. & Prompt or Source chunking  \\
\midrule
SQC$\downarrow$ RQC$\uparrow$ GR$\downarrow$ & Generator responding to queries (or part thereof) that are not addressed in source, causing ungroundedness.  & Prompt \\
\bottomrule
\end{tabularx}
\caption{Diagnosis of RAG QA applications with interplay of THELMA metrics. Arrows denote low($\downarrow$) and high($\uparrow$) metric scores.}
\label{tab:interpretation}
\end{table*}
We augment and report results on randomly sampled subset ($240$ data points) of WikiEval from RAGAs \citep{es-etal-2024-ragas}. WikiEval dataset is augmented manually by modifying data points to contain sub-optimal sources or responses representing issues assessed by each metrics. The dataset is annotated with preferential data annotation strategy. For each instance in the dataset, annotators were given two sets of source or response pairs for every query, $q$. All annotators are developers or scientists from the adopter teams and are fluent in English.They were given clear instructions with metrics definitions. We then compared agreement of RAGAs and THELMA with human annotation in pairwise comparison. An agreement between human and evaluation framework would entail evaluation framework scoring the better source, response higher than the other. For fair comparison of frameworks, we use the same underlying LLM for both frameworks and compare our metrics against the most relevant metrics from RAGAs. Our proposed metrics consistently score the better source set and response with higher value (see Table \ref{tab:response-metrics-results}). Consistent to conclusions by \citet{es-etal-2024-ragas}, we find the source metrics to be the hardest dimensions to evaluate. Open source model(Llama) based THELMA also produced comparable results to proprietary models. Table \ref{tab:interpretation} demonstrate how THELMA scores are interpreted by application developers to make further improvements.
\section{Conclusion}
This paper presents THELMA, a novel reference free evaluation framework for RAG based question answering task. The framework provides granular independent assessment of retrievers and generators while also providing insights to application developers on specific areas of improvements by introducing precision-coverage trade off. The evaluation of the framework on THELMA-WikiEval dataset has shown consistent alignment with human judgments regardless of the underlying LLM.
\section{Limitations}
The framework is unlikely to be useful to evaluate QA systems with unverified source corpus. Secondly, the framework does not perform well on creative or opinionated questions e.g. summarization, essay writing, text comparisons. In the current state, the framework has only been evaluated for its performance on unstructured text sources. Thirdly, the framework heavily depends on the performance of underlying LLMs. 


\section*{Ethics Statement}
We do not anticipate any ethical concerns with the framework presented here. In terms of cost, proposed metrics produce comparable performance with open source LLMs. Customer data or personally identifiable information is not discussed or released through this paper.   

\label{sec:bibtex}

\bibliography{THELMA}

\begin{thebibliography}{22}
\providecommand{\natexlab}[1]{#1}

\bibitem[{Briakou et~al.(2024)Briakou, Liu, Cherry, and Freitag}]{briakou2024implicationsverbosellmoutputs}
Eleftheria Briakou, Zhongtao Liu, Colin Cherry, and Markus Freitag. 2024.
\newblock \href {https://arxiv.org/abs/2410.00863} {On the implications of verbose llm outputs: A case study in translation evaluation}.
\newblock \emph{Preprint}, arXiv:2410.00863.

\bibitem[{Chen et~al.(2023)Chen, Lin, Han, and Sun}]{chen2023benchmarkinglargelanguagemodels}
Jiawei Chen, Hongyu Lin, Xianpei Han, and Le~Sun. 2023.
\newblock \href {https://arxiv.org/abs/2309.01431} {Benchmarking large language models in retrieval-augmented generation}.
\newblock \emph{Preprint}, arXiv:2309.01431.

\bibitem[{Cuconasu et~al.(2024)Cuconasu, Trappolini, Siciliano, Filice, Campagnano, Maarek, Tonellotto, and Silvestri}]{10.1145/3626772.3657834}
Florin Cuconasu, Giovanni Trappolini, Federico Siciliano, Simone Filice, Cesare Campagnano, Yoelle Maarek, Nicola Tonellotto, and Fabrizio Silvestri. 2024.
\newblock \href {https://doi.org/10.1145/3626772.3657834} {The power of noise: Redefining retrieval for rag systems}.
\newblock In \emph{Proceedings of the 47th International ACM SIGIR Conference on Research and Development in Information Retrieval}, SIGIR '24, page 719–729, New York, NY, USA. Association for Computing Machinery.

\bibitem[{Es et~al.(2024)Es, James, Espinosa~Anke, and Schockaert}]{es-etal-2024-ragas}
Shahul Es, Jithin James, Luis Espinosa~Anke, and Steven Schockaert. 2024.
\newblock \href {https://aclanthology.org/2024.eacl-demo.16} {{RAGA}s: Automated evaluation of retrieval augmented generation}.
\newblock In \emph{Proceedings of the 18th Conference of the European Chapter of the Association for Computational Linguistics: System Demonstrations}, pages 150--158, St. Julians, Malta. Association for Computational Linguistics.

\bibitem[{Hu et~al.(2024)Hu, Ru, Qiu, Guo, Zhang, Xu, Luo, Liu, Zhang, and Zhang}]{hu_refchecker_2024}
Xiangkun Hu, Dongyu Ru, Lin Qiu, Qipeng Guo, Tianhang Zhang, Yang Xu, Yun Luo, Pengfei Liu, Yue Zhang, and Zheng Zhang. 2024.
\newblock \href {http://arxiv.org/abs/2405.14486} {{RefChecker}: {Reference}-based {Fine}-grained {Hallucination} {Checker} and {Benchmark} for {Large} {Language} {Models}}.
\newblock \emph{arXiv preprint}.
\newblock ArXiv:2405.14486 [cs].

\bibitem[{Huang et~al.(2025)Huang, Yu, Ma, Zhong, Feng, Wang, Chen, Peng, Feng, Qin, and Liu}]{Huang_2025}
Lei Huang, Weijiang Yu, Weitao Ma, Weihong Zhong, Zhangyin Feng, Haotian Wang, Qianglong Chen, Weihua Peng, Xiaocheng Feng, Bing Qin, and Ting Liu. 2025.
\newblock \href {https://doi.org/10.1145/3703155} {A survey on hallucination in large language models: Principles, taxonomy, challenges, and open questions}.
\newblock \emph{ACM Transactions on Information Systems}, 43(2):1–55.

\bibitem[{J.~Ferrara(2024)}]{trulensTriad}
O.~M.~Ozturk J.~Ferrara, Ethan-Tonic. 2024.
\newblock Rag triad; {T}ru{L}ens trulens.org.
\newblock \url{https://www.trulens.org/getting_started/core_concepts/rag_triad/}.
\newblock [Accessed 11-03-2025].

\bibitem[{Lewis et~al.(2020)Lewis, Perez, Piktus, Petroni, Karpukhin, Goyal, K\"{u}ttler, Lewis, Yih, Rockt\"{a}schel, Riedel, and Kiela}]{NEURIPS2020_6b493230}
Patrick Lewis, Ethan Perez, Aleksandra Piktus, Fabio Petroni, Vladimir Karpukhin, Naman Goyal, Heinrich K\"{u}ttler, Mike Lewis, Wen-tau Yih, Tim Rockt\"{a}schel, Sebastian Riedel, and Douwe Kiela. 2020.
\newblock \href {https://proceedings.neurips.cc/paper_files/paper/2020/file/6b493230205f780e1bc26945df7481e5-Paper.pdf} {Retrieval-augmented generation for knowledge-intensive nlp tasks}.
\newblock In \emph{Advances in Neural Information Processing Systems}, volume~33, pages 9459--9474. Curran Associates, Inc.

\bibitem[{Liu et~al.(2024)Liu, Lin, Hewitt, Paranjape, Bevilacqua, Petroni, and Liang}]{liu-etal-2024-lost}
Nelson~F. Liu, Kevin Lin, John Hewitt, Ashwin Paranjape, Michele Bevilacqua, Fabio Petroni, and Percy Liang. 2024.
\newblock \href {https://doi.org/10.1162/tacl_a_00638} {Lost in the middle: How language models use long contexts}.
\newblock \emph{Transactions of the Association for Computational Linguistics}, 12:157--173.

\bibitem[{Min et~al.(2023)Min, Krishna, Lyu, Lewis, Yih, Koh, Iyyer, Zettlemoyer, and Hajishirzi}]{min-etal-2023-factscore}
Sewon Min, Kalpesh Krishna, Xinxi Lyu, Mike Lewis, Wen-tau Yih, Pang Koh, Mohit Iyyer, Luke Zettlemoyer, and Hannaneh Hajishirzi. 2023.
\newblock \href {https://doi.org/10.18653/v1/2023.emnlp-main.741} {{FA}ct{S}core: Fine-grained atomic evaluation of factual precision in long form text generation}.
\newblock In \emph{Proceedings of the 2023 Conference on Empirical Methods in Natural Language Processing}, pages 12076--12100, Singapore. Association for Computational Linguistics.

\bibitem[{Ru et~al.(2024{\natexlab{a}})Ru, Qiu, Hu, Zhang, Shi, Chang, Jiayang, Wang, Sun, Li, Zhang, Wang, Jiang, He, Wang, Liu, Zhang, and Zhang}]{ru2024ragcheckerfinegrainedframeworkdiagnosing}
Dongyu Ru, Lin Qiu, Xiangkun Hu, Tianhang Zhang, Peng Shi, Shuaichen Chang, Cheng Jiayang, Cunxiang Wang, Shichao Sun, Huanyu Li, Zizhao Zhang, Binjie Wang, Jiarong Jiang, Tong He, Zhiguo Wang, Pengfei Liu, Yue Zhang, and Zheng Zhang. 2024{\natexlab{a}}.
\newblock \href {https://arxiv.org/abs/2408.08067} {Ragchecker: A fine-grained framework for diagnosing retrieval-augmented generation}.
\newblock \emph{Preprint}, arXiv:2408.08067.

\bibitem[{Ru et~al.(2024{\natexlab{b}})Ru, Qiu, Hu, Zhang, Shi, Chang, Jiayang, Wang, Sun, Li, Zhang, Wang, Jiang, He, Wang, Liu, Zhang, and Zhang}]{ru_ragchecker_2024}
Dongyu Ru, Lin Qiu, Xiangkun Hu, Tianhang Zhang, Peng Shi, Shuaichen Chang, Cheng Jiayang, Cunxiang Wang, Shichao Sun, Huanyu Li, Zizhao Zhang, Binjie Wang, Jiarong Jiang, Tong He, Zhiguo Wang, Pengfei Liu, Yue Zhang, and Zheng Zhang. 2024{\natexlab{b}}.
\newblock \href {http://arxiv.org/abs/2408.08067} {{RAGChecker}: {A} {Fine}-grained {Framework} for {Diagnosing} {Retrieval}-{Augmented} {Generation}}.
\newblock \emph{arXiv preprint}.
\newblock ArXiv:2408.08067 [cs].

\bibitem[{Saad-Falcon et~al.(2024)Saad-Falcon, Khattab, Potts, and Zaharia}]{saad-falcon-etal-2024-ares}
Jon Saad-Falcon, Omar Khattab, Christopher Potts, and Matei Zaharia. 2024.
\newblock \href {https://doi.org/10.18653/v1/2024.naacl-long.20} {{ARES}: An automated evaluation framework for retrieval-augmented generation systems}.
\newblock In \emph{Proceedings of the 2024 Conference of the North American Chapter of the Association for Computational Linguistics: Human Language Technologies (Volume 1: Long Papers)}, pages 338--354, Mexico City, Mexico. Association for Computational Linguistics.

\bibitem[{Salemi and Zamani(2024)}]{salemi2024evaluating}
Alireza Salemi and Hamed Zamani. 2024.
\newblock Evaluating retrieval quality in retrieval-augmented generation.
\newblock In \emph{Proceedings of the 47th International ACM SIGIR Conference on Research and Development in Information Retrieval}, pages 2395--2400.

\bibitem[{Voorhees and Tice(2000)}]{voorhees-tice-2000-trec}
Ellen~M. Voorhees and Dawn~M. Tice. 2000.
\newblock \href {https://aclanthology.org/L00-1018/} {The {TREC}-8 question answering track}.
\newblock In \emph{Proceedings of the Second International Conference on Language Resources and Evaluation ({LREC}`00)}, Athens, Greece. European Language Resources Association (ELRA).

\bibitem[{Wang et~al.(2024)Wang, Hernandez, Kyslyi, and Kersting}]{wang2024evaluatingqualityanswersretrievalaugmented}
Yang Wang, Alberto~Garcia Hernandez, Roman Kyslyi, and Nicholas Kersting. 2024.
\newblock \href {https://arxiv.org/abs/2406.18064} {Evaluating quality of answers for retrieval-augmented generation: A strong llm is all you need}.
\newblock \emph{Preprint}, arXiv:2406.18064.

\bibitem[{Wu et~al.(2024)Wu, Xie, Chen, Zhu, Zhang, and Xiao}]{wu2024how}
Siye Wu, Jian Xie, Jiangjie Chen, Tinghui Zhu, Kai Zhang, and Yanghua Xiao. 2024.
\newblock \href {https://openreview.net/forum?id=S7NVVfuRv8} {How easily do irrelevant inputs skew the responses of large language models?}
\newblock In \emph{First Conference on Language Modeling}.

\bibitem[{Xu et~al.(2024)Xu, Cruz, Guevara, Wang, Deshpande, Wang, and Li}]{10.1145/3626772.3661370}
Zhentao Xu, Mark~Jerome Cruz, Matthew Guevara, Tie Wang, Manasi Deshpande, Xiaofeng Wang, and Zheng Li. 2024.
\newblock \href {https://doi.org/10.1145/3626772.3661370} {Retrieval-augmented generation with knowledge graphs for customer service question answering}.
\newblock In \emph{Proceedings of the 47th International ACM SIGIR Conference on Research and Development in Information Retrieval}, SIGIR '24, page 2905–2909, New York, NY, USA. Association for Computing Machinery.

\bibitem[{Zhang et~al.(2024{\natexlab{a}})Zhang, Jiang, Bai, Zhang, Lin, Liu, and Ren}]{zhang-etal-2024-rag4itops}
Tianyang Zhang, Zhuoxuan Jiang, Shengguang Bai, Tianrui Zhang, Lin Lin, Yang Liu, and Jiawei Ren. 2024{\natexlab{a}}.
\newblock \href {https://doi.org/10.18653/v1/2024.emnlp-industry.56} {{RAG}4{ITO}ps: A supervised fine-tunable and comprehensive {RAG} framework for {IT} operations and maintenance}.
\newblock In \emph{Proceedings of the 2024 Conference on Empirical Methods in Natural Language Processing: Industry Track}, pages 738--754, Miami, Florida, US. Association for Computational Linguistics.

\bibitem[{Zhang et~al.(2020)Zhang, Kishore, Wu, Weinberger, and Artzi}]{zhang2020bertscoreevaluatingtextgeneration}
Tianyi Zhang, Varsha Kishore, Felix Wu, Kilian~Q. Weinberger, and Yoav Artzi. 2020.
\newblock \href {https://arxiv.org/abs/1904.09675} {Bertscore: Evaluating text generation with bert}.
\newblock \emph{Preprint}, arXiv:1904.09675.

\bibitem[{Zhang et~al.(2024{\natexlab{b}})Zhang, Das, and Zhang}]{zhang2024verbosityneqveracitydemystify}
Yusen Zhang, Sarkar Snigdha~Sarathi Das, and Rui Zhang. 2024{\natexlab{b}}.
\newblock \href {https://arxiv.org/abs/2411.07858} {Verbosity $\neq$ veracity: Demystify verbosity compensation behavior of large language models}.
\newblock \emph{Preprint}, arXiv:2411.07858.

\bibitem[{Zheng et~al.(2024)Zheng, Chiang, Sheng, Zhuang, Wu, Zhuang, Lin, Li, Li, Xing, Zhang, Gonzalez, and Stoica}]{10.5555/3666122.3668142}
Lianmin Zheng, Wei-Lin Chiang, Ying Sheng, Siyuan Zhuang, Zhanghao Wu, Yonghao Zhuang, Zi~Lin, Zhuohan Li, Dacheng Li, Eric~P. Xing, Hao Zhang, Joseph~E. Gonzalez, and Ion Stoica. 2024.
\newblock Judging llm-as-a-judge with mt-bench and chatbot arena.
\newblock In \emph{Proceedings of the 37th International Conference on Neural Information Processing Systems}, NIPS '23, Red Hook, NY, USA. Curran Associates Inc.

\end{thebibliography}

\appendix
\label{sec:appendix}
\section{THELMA Examples}
\def\code#1{\texttt{#1}}
\subsection{Source precision}
\begin{enumerate}
\item \code{\textbf{Query}: When was the Chimnabai Clock Tower completed, and who was it named after?}
\item \code{\textbf{Sources and match scores }}
\begin{enumerate}
\item\code{The Chimnabai Clock Tower, also known as the Raopura Tower, is a clock tower situated in the Raopura area of Vadodara, Gujarat, India. It was completed in 1896 and named in memory of Chimnabai I (1864‚Äì1885), a queen and the first wife of Sayajirao Gaekwad III of Baroda State. It was built in Indo-Saracenic architecture style. $score=1$}
\item\code{ Chimnabai Clock Tower was inaugurated by Mir Kamaluddin Hussainkhan, the last Nawab of Baroda. During the rule of Gaekwad, it was a stoppage for horse drawn trams. It was constructed from the funds raised by the citizens of Baroda State.The premises of the tower were rented by the Vadodara Municipal Corporation for the last three decades. In 2017, the municipal commissioner, Vinod Rao, ordered to vacate the tower to conserve it as a heritage monument. The mechanical system of the tower clock was replaced by DC motor.' $score=0$}
\end{enumerate}
\item \code{\texttt{\textbf{Source Precision}: $0.5$}}
\end{enumerate}

\subsection{Source Query Coverage}
\begin{enumerate}
\item \code{\textbf{Source}: The Chimnabai Clock Tower, also known as the Raopura Tower, is a clock tower situated in the Raopura area of Vadodara, Gujarat, India. It was completed in 1896. It was built in Indo-Saracenic architecture style.\\ History\\ It was inaugurated by Mir Kamaluddin Hussainkhan, the last Nawab of Baroda. During the rule of Gaekwad, it was a stoppage for horse drawn trams. It was constructed from the funds raised by the citizens of Baroda State.The premises of the tower were rented by the Vadodara Municipal Corporation for the last three decades. In 2017, the municipal commissioner, Vinod Rao, ordered to vacate the tower to conserve it as a heritage monument. The mechanical system of the tower clock was replaced by DC motor.'}
\item \code{\textbf{Query}: When was the Chimnabai Clock Tower completed, and who was it named after?}
\item \code{\textbf{Decomposed query and match scores :}}
\begin{enumerate}
    \item \code{When was the Chimnabai Clock Tower completed? $score = 1$}
    \item \code{Who was Chimnabai Clock Tower named after? $score = 0$}
\end{enumerate}
\item \code \texttt{\textbf{Source Query Coverage}: $0.5$}
\end{enumerate}
\subsection{Response Precision}
\begin{enumerate}
\item \code{\textbf{Query}: When was the Chimnabai Clock Tower completed, and who was it named after?
\item\textbf{Response}:The Chimnabai Clock Tower was completed in 1896. It was named after Chimnabai I, who was a queen and the first wife of Sayajirao Gaekwad III of Baroda State. The construction of clock tower was completed in 1896. 
\item \textbf{Decomposes response and match scores:}
\begin{enumerate}
    \item \code{The Chimnabai Clock Tower was completed in 1896. $score=1$}
    \item \code{The place was named after Chimnabai I. $score=1$}
    \item \code{Chimnabai I was a queen. $score=0$}
    \item \code{Chimnabai I was the first wife of Sayajirao Gaekwad III. $score=0$}
    \item \code{Sayajirao Gaekwad III was the ruler of Baroda State. $score=0$}
    \item \code{The clock tower was constructed. $score=0$}
    \item \code{The construction of the clock tower was completed in 1896. $score=1$}
\end{enumerate}}
\item \texttt{\textbf{Response Precision}: $0.42$}
\end{enumerate}
\subsection{Response Query Coverage}
\begin{enumerate}
\item \code{\textbf{Query}: When was the Chimnabai Clock Tower completed, and who was it named after?}
\item \code{\textbf{Response}:The Chimnabai Clock Tower was completed in 1896.}
\item \code{\textbf{Decomposed query and match scores :}}
\begin{enumerate}
    \item \code{When was the Chimnabai Clock Tower completed? $score = 1$}
    \item \code{Who was Chimnabai Clock Tower named after? $score = 0$}
\end{enumerate}

\item \code \texttt{\textbf{Response Query Coverage}: $0.5$}
\end{enumerate}

\subsection{Response Self Distinctness}
\begin{enumerate}
\item \code{\textbf{Response}:The Chimnabai Clock Tower was completed in 1896. It was named after Chimnabai I, who was a queen and the first wife of Sayajirao Gaekwad III of Baroda State. The construction of clock tower was completed in 1896. }
\item \code{\textbf{Decomposes response and match scores:}}
\begin{enumerate}
    \item \code{The Chimnabai Clock Tower was completed in 1896. $score=1$}
    \item \code{It was named after Chimnabai I, who was a queen and the first wife of Sayajirao Gaekwad III of Baroda State. $score=0$}
    \item \code{The construction of clock tower was completed in 1896. $score=1$}
\end{enumerate}
\item \texttt{\textbf{Response Self Distinctness}: $1-0.66 = 0.33$}
\end{enumerate}
\subsection{Groundedness}
\begin{enumerate}
\item \code{\textbf{Source}: The Chimnabai Clock Tower, also known as the Raopura Tower, is a clock tower situated in the Raopura area of Vadodara, Gujarat, India. It was completed in 1896 and named in memory of Chimnabai I (1864‚Äì1885), a queen and the first wife of Sayajirao Gaekwad III of Baroda State. It was built in Indo-Saracenic architecture style.\\ History\\ Chimnabai Clock Tower was built in 1896. The tower was named after Chimnabai I (1864‚Äì1885), a queen and the first wife of Sayajirao Gaekwad III of Baroda State. It was inaugurated by Mir Kamaluddin Hussainkhan, the last Nawab of Baroda. During the rule of Gaekwad, it was a stoppage for horse drawn trams. It was constructed from the funds raised by the citizens of Baroda State.The premises of the tower were rented by the Vadodara Municipal Corporation for the last three decades. In 2017, the municipal commissioner, Vinod Rao, ordered to vacate the tower to conserve it as a heritage monument. The mechanical system of the tower clock was replaced by DC motor.'}
\item\code{\textbf{Response}:The Chimnabai Clock Tower was completed in 1856. It was named after Chimnabai I, who was a queen and the first wife of Sayajirao Gaekwad III of Baroda State. The construction of clock tower was completed in 1856. }
\item \code{\textbf{Decomposes response and match scores:}}
\begin{enumerate}
    \item \code{The Chimnabai Clock Tower was completed in 1896. $score=0$}
    \item \code{The place was named after Chimnabai I. $score=1$}
    \item \code{Chimnabai I was a queen. $score=1$}
    \item \code{Chimnabai I was the first wife of Sayajirao Gaekwad III. $score=1$}
    \item \code{Sayajirao Gaekwad III was the ruler of Baroda State. $score=1$}
    \item \code{The clock tower was constructed. $score=1$}
    \item \code{The construction of the clock tower was completed in 1896. $score=0$}
\end{enumerate}
\item \texttt{\textbf{Groundedness}: $0.71$}
\end{enumerate}
\section{Prompts}
\subsection{Source, Response Decompose Module}
\code{You are a claim extractor and your job is to extract claims from snippets of text. You always follow the below guidelines:
\\\\
<guidelines> \\}
\code{
- You only respond with a list of stand-alone claims. \\
- The list is always inside an <output> </output> tags. \\
- The extracted claims contain only one piece of stand-alone information such that each claim can be verified independently. \\
- The extracted claims should be independently verifiable but do not need to be true. \\
- The extracted claims should be stand-alone facts. \\ 
- The extracted claims are independent so there is very little overlap between them. \\
- The list of extracted claims should contain all information from the input. \\
- Each extracted claim should not be decomposable into multiple claims. \\
- Generate as many claims as possible. \\
</guidelines> \\}
\\
\\
\code{
Human: Here are some examples: \\
<examples> \\
Human: Please extract claims from the following text: \\
<input> \\
Text: Jay got his PhD in Santa Barbara and during that time he got a dog named Basil. His PhD was focused on the mathematics of plasmas and he did some deep learning on the side.
</input> \\\\
}
\code{Assistant:\\
<output> \\
- Jay got a PhD in Santa Barbara. \\
- Jay got a dog named Basil in Santa Barbara. \\
- Basil is a dog from Santa Barbara. \\
- Jay worked on plasmas during his PhD. \\
- Jay worked on deep learning during his PhD. \\
</output> \\}
\code{
</examples> \\\\
Human: Please extract claims from the following text:
<input>
\{input\}
</input>}

\subsection{Response Precision and Source Precision Match Module}
\code{You are an editor. Your task is to identify if given fact contains essential information to answer a query. You will be given a pair of query and fact. Your task is to rate whether the fact is essential to answer the query.  Please follow the instructions carefully.} \\ \\
\code{Evaluation Criteria:
Relevance Score captures whether the information in the fact is essential to answer the specific query asked.
This dimension assesses if the response provides relevant details and does not have any irrelevant details. \\
} \\
\code{Instructions: \\
<instructions> \\
    - Read the query carefully to thoroughly understand the user's query. Identify the key points that are being asked about. \\
    - Analyze the response to ensure it is essential to answer the query. \\
    - Check if response is providing details about same entity or event asked in the query. \\
    - Check if the response provides any extraneous details which are not needed to answer the query. \\
    - Provide the response in <output> </output> tags. \\
    - Respond only with a value in the scale provided below. No explanations.\\
    </instructions> \\ \\
}
\code{Scale: \\ 
<scale> \\
Extraneous - The response is not required to answer the core intent of the query. \\
Essential - The response is essential to answer the query. \\
</scale> \\
}
\code{Human: Please rate the essentiality of the fact to answer the query: \\
<input> \\
Question: \{query\} \\
Response: \{response\} \\
</input>} \\

\subsection{Query Decompose Module}
\code{You are a question decomposer. You will follow following guidelines. \\
\\
<guidelines> \\
- You only respond with a list of questions. \\
- The list is always inside of <output> </output> tags. \\
- You should resolve any ambiguous pronouns in the input text. \\
- Each extracted question contain only one question. \\
- Each extracted question is one short question. \\
- All the extracted questions are from the input questions received. \\
</guidelines> \\
}
\code{Here are some examples: \\
<examples> \\
Human: Please extract questions from the following text: \\
<input> \\
Text: Imagine you are a travel agent and you are trying to book flights to London. What is the most effective way to book flights to London?
</input> \\
}
\code{Assistant: <output>
What is the most effective way to book flights to London? \\
</output> \\
</examples> \\
Human: Please extract questions from the following text: \\
<input> \\
Text: \{input\} \\
</input>}

\subsection{Groundedness Match Module}

\code{
CLAIM\_EXAMPLE\_KS\_1 \= Hawaii is allegedly named after Hawai'iloa, a legendary Polynesian navigator who is said to have discovered the island. Other accounts attribute the name to the legendary realm of Hawaiki, a place from which some Polynesians are said to have originated, the place where they transition to in the afterlife, or the realm of the gods and goddesses. James Cook, the English explorer and navigator who captained the first European expedition to reach the Hawaiian Islands, called it O-Why-hee (from Hawaiian) and the Sandwich Islands after his patron, the Earl of Sandwich.} \\ \\
\code{CLAIM\_EXAMPLE\_KS\_2 \=
- You can deploy a model trained with SageMaker to your own deployment target. To do that, you need to know the algorithm-specific format of the model artifacts that were generated by model training. For more information about output formats, see the section corresponding to the algorithm you are using in Common Data Formats for Training.
- You can deploy multiple variants of a model to the same SageMaker HTTPS endpoint. This is useful for testing variations of a model in production. For example, suppose that you've deployed a model into production. You want to test a variation of the model by directing a small amount of traffic, say 5\%, to the new model. To do this, create an endpoint configuration that describes both variants of the model. You specify the ProductionVariant in your request to the CreateEndPointConfig. For more information, see ProductionVariant. \\\\} 
\code{You are a fact checker. You will be given a claim and a knowledge base. You will then answer with a 0 or 1 indicating if the claim is not support or supported by the knowledge base. The scale is addressed here: \\
<scale> \\
- 0: The claim is not supported by the knowledge base. \\
- 1: The claim is definitely supported by the knowledge base. \\
</scale> \\}
\code{
You always adhere to the following guidelines: \\
<guidelines> \\
- The output is always between <output> and </output>. \\
- Only the score is returned. Explanations are forbidden. \\
</guidelines> \\
}
\code{
Here is an example: \\
<example> \\
<example1> \\
Human: Is the claim supported by the knowledge base? \\
<input> \\
}
\code{
<knowledge source>
\{CLAIM\_EXAMPLE\_KS\_1\} \\
</knowledge source> \\
<claim> \\
The capital of Hawaii is Honolulu. \\
</claim> \\
</input> \\
}
\code{Assistant: <output>0</output> \\
</example1> \\\\
<example2> \\
Human: Is the claim supported by the knowledge base? \\
<input> \\
<knowledge source> \\
\{CLAIM\_EXAMPLE\_KS\_2\} \\
</knowledge source> \\
<claim> \\
SageMaker allows you to deploy different model variants to the same endpoint. \\
</claim> \\
</input> \\ 
Assistant: <output>1</output> \\
</example2> \\
</example> \\
Human: Is the claim supported by the knowledge base? \\
}

\section{Human Annotation}
As mentioned in Section \ref{sec:experiments}, we manually augmented on a randomly sampled subset of WikiEval dataset from RAGAs \citep{es-etal-2024-ragas}. Originally, each data point in this dataset contains a question, correct response (positive), correct sources (positive), ungrounded response (negative), irrelavant response (negative) and irrelevant sources (negative). We modified this dataset by adding sub-optimal responses and sources for each datapoint corresponding to each metric. Specifically, we took a subset of 20 datapoints from WikiEval dataset and added the following columns for each of them.
\begin{itemize}
    \item We took the correct response and added irrelevant facts that is not essential to answer the question. This serves as negative for Response Precision metric i.e., We test whether the Response Precision metric scores the correct response higher than the modified negative response.
    \item We took the correct response and removed one of the essential piece of information to answer the question. This serves as negative for Response Query Coverage metric i.e, We test whether the Response Query Coverage metric scores the correct response higher than the one with missing essential points.
    \item We took the correct response and rephrased the facts in the response and added them to the response such that it contains redundant information. We use this response with redundant information as negative to test Self-distinctness whether the metric scores correct response higher than the response with redundant information.
    \item We took the correct set of sources and added randomly chosen irrelevant sources to them to create negative for Source Precision metric at chunk level ( variant 1 ). For fact level variants, we added irrelevant information to some of the sources to create the negatives. Using these negatives, we test whether the Source Precision metric scores the correct sources higher than the negative sources.
    \item We took the correct set of sources and removed essential facts necessary to answer the question. This serves as negative for Source Query Coverage netric which we used to test whether Source Query Coverage metric scores the correct sources higher than the  sources with missing essential information.
\end{itemize}

In this way, we create negatives for each dimension and compare our metric with RAGAs. As mentioned in Section \ref{sec:related-work}, RAGAs does not contain coarse-grained metrics. Hence, we compare THELMA metrics with its closest counterpart in RAGAs as mentioned in \ref{tab:response-metrics-results}.
\end{document}